\documentclass[sigconf]{acmart}

\usepackage{graphicx}  
\usepackage{xspace}
\usepackage{eufrak}
\usepackage{bm}
\usepackage{multirow}
\usepackage{booktabs}
\usepackage{subfig}
\usepackage[utf8]{inputenc}

\usepackage{cleveref}
\usepackage{enumitem}
\usepackage{underscore}
\usepackage{adjustbox}

\usepackage{xargs}
\usepackage{etoolbox}
\usepackage[flushleft]{threeparttable}

\usepackage[colorinlistoftodos,prependcaption,textsize=tiny]{todonotes}
\usepackage[group-separator={,}, group-minimum-digits=4]{siunitx}

\setlist{nosep}

\newcommand{\safedefine}[2]{
  \ifdef{#1}{\renewcommand{#1}{#2}}{\newcommand{#1}{#2}}
}
\makeatletter
\newcommand\footnoteref[1]{\protected@xdef\@thefnmark{\ref{#1}}\@footnotemark}
\makeatother

\newcommand{\ours}{{Sparse-CHMM}\xspace}

\newcommand*{\rom}[1]{\uppercase\expandafter{\romannumeral #1}}

\newcommandx{\chao}[2][1=]{\todo[linecolor=red,backgroundcolor=red!25,bordercolor=red,#1]{#2}}
\newcommandx{\yli}[2][1=]{\todo[linecolor=blue,backgroundcolor=blue!25,bordercolor=blue,#1]{#2}}
\newcommandx{\lsong}[2][1=]{\todo[linecolor=green,backgroundcolor=green!25,bordercolor=green,#1]{#2}}
\newcommandx{\pranav}[2][1=]{\todo[linecolor=purple,backgroundcolor=purple!25,bordercolor=red,#1]{#2}}
\newcommandx{\improvement}[2][1=]{\todo[linecolor=yellow,backgroundcolor=yellow!25,bordercolor=yellow,#1]{#2}}
\newcommandx{\thiswillnotshow}[2][1=]{\todo[disable,#1]{#2}}

\newcommand{\lbfont}[1]{\texttt{#1}}

\newcommand{\ie}{\emph{i.e.}\xspace} 
\newcommand{\wrt}{\emph{w.r.t.}\xspace}
\newcommand{\aka}{\emph{a.k.a.}\xspace}
\safedefine{\iff}{\emph{i.f.f.}\xspace}
\safedefine{\iid}{\emph{i.i.d.}\xspace}
\newcommand{\fone}{F\textsubscript{1}\xspace}
\newcommand{\vs}{\emph{vs.}\xspace}

\DeclareMathOperator*{\argmax}{arg\,max}

\newcommand{\A}{\bm{A}}

\safedefine{\C}{\bm{C}}
\newcommand{\W}{\bm{W}}
\newcommand{\x}{\bm{x}}
\newcommand{\0}{\bm{0}}
\newcommand{\y}{\bm{y}}

\newcommand{\z}{\bm{z}}
\newcommand{\e}{\bm{e}}
\newcommand{\w}{\bm{w}}

\newcommand{\ents}{\mathcal{E}}
\newcommand{\ent}{{\rm ent}}
\newcommand{\lbs}{\mathcal{L}}
\newcommand{\R}{\mathbb{R}}

\safedefine{\I}{\mathbb{I}}

\safedefine{\O}{\mathcal{O}}

\newcommand{\bPsi}{\bm{\Psi}}
\newcommand{\bPhi}{\bm{\Phi}}
\newcommand{\bphi}{\bm{\varphi}}
\newcommand{\balpha}{\bm{\alpha}}
\newcommand{\bbeta}{\bm{\beta}}

\newcommand{\bgamma}{\bm{\gamma}}

\newcommand{\btht}{\bm{\theta}}
\newcommand{\bLambda}{\bm{\Lambda}}
\newcommand{\bDelta}{\bm{\Delta}}
\newcommand{\bOmega}{\bm{\Omega}}
\newcommand{\bxi}{\bm{\xi}}

\newcommand{\expt}{\mathbb{E}}
\newcommand{\softmax}{\text{SoftMax}}

\newcommand{\fc}{\text{FC}}

\newcommand{\lquery}{l^{\rm query}}
\newcommand{\ltgt}{l^{\rm tgt}}
\safedefine{\Dir}{\text{Dir}}

\AtBeginDocument{%
  \providecommand\BibTeX{{%
    \normalfont B\kern-0.5em{\scshape i\kern-0.25em b}\kern-0.8em\TeX}}}

\copyrightyear{2022}
\acmYear{2022}
\setcopyright{rightsretained}
\acmConference[KDD '22]{Proceedings of the 28th ACM SIGKDD Conference on Knowledge Discovery and Data Mining}{August 14--18, 2022}{Washington, DC, USA}
\acmBooktitle{Proceedings of the 28th ACM SIGKDD Conference on Knowledge Discovery and Data Mining (KDD '22), August 14--18, 2022, Washington, DC, USA}
\acmISBN{978-1-4503-9385-0/22/08}
\acmDOI{10.1145/3534678.3539247}

\usepackage{etoolbox}
\makeatletter
\patchcmd{\maketitle}{\@copyrightpermission}{
   \begin{minipage}{0.3\columnwidth}
     \href{https://creativecommons.org/licenses/by/4.0/}{\includegraphics[width=0.90\textwidth]{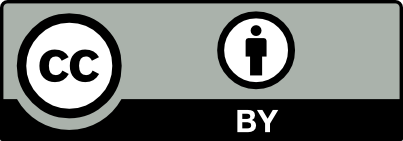}}
   \end{minipage}\hfill
   \begin{minipage}{0.7\columnwidth}
     \href{https://creativecommons.org/licenses/by/4.0/}{This work is licensed under a Creative Commons Attribution International 4.0 License.}
   \end{minipage}
  
   \vspace{5pt}
}{}{}

\makeatother

\acmSubmissionID{rtfp0298}

\begin{document}

\title{Sparse Conditional Hidden Markov Model for Weakly Supervised Named Entity Recognition}

\author{Yinghao Li}
\orcid{0000-0002-7188-4136}
\affiliation{%
  \institution{Georgia Institute of Technology}
  \city{Atlanta}
  \state{Georgia}
  \country{USA}
}
\email{yinghaoli@gatech.edu}

\author{Le Song}
\orcid{0000-0003-0345-9899}
\affiliation{%
  \institution{BioMap, MBZUAI}
  \city{Beijing}
  \country{China}
}
\email{le.song@mbzuai.ac.ae}

\author{Chao Zhang}
\orcid{0000-0003-3009-598X}
\affiliation{%
  \institution{Georgia Institute of Technology}
  \city{Atlanta}
  \state{Georgia}
  \country{USA}
}
\email{chaozhang@gatech.edu}

\renewcommand{\shortauthors}{Yinghao Li, Le Song, \& Chao Zhang}


\begin{abstract}
  Weakly supervised named entity recognition methods train \emph{label models} to aggregate the token annotations of multiple noisy labeling functions (LFs) without seeing any manually annotated labels.
  To work well, the label model needs to contextually identify and emphasize well-performed LFs while down-weighting the under-performers.
  However, evaluating the LFs is challenging due to the lack of ground truths.
  To address this issue, we propose the \emph{sparse conditional hidden Markov model} (\ours).
  Instead of predicting the entire emission matrix as other HMM-based methods, \ours focuses on estimating its diagonal elements, which are considered as the reliability scores of the LFs.
  The sparse scores are then expanded to the full-fledged emission matrix with pre-defined expansion functions.
  We also augment the emission with weighted XOR scores, which track the probabilities of an LF observing incorrect entities.
  \ours is optimized through unsupervised learning with a three-stage training pipeline that reduces the training difficulty and prevents the model from falling into local optima.
  Compared with the baselines in the Wrench benchmark, \ours achieves a $3.01$ average \fone score improvement on \num{5} comprehensive datasets.
  Experiments show that each component of \ours is effective, and the estimated LF reliabilities strongly correlate with true LF \fone scores.
\end{abstract}



\maketitle

\section{Introduction}
\label{sec:introduction}

Named entity recognition (NER), a task that identifies pre-defined named entities, is fundamental for extracting information from unstructured text data.
For example, to facilitate material synthesis, material scientists often need to extract entities such as materials, properties, and catalysts from research articles \citep{kim.2017.Materials.Synthesis}.
NER also powers tasks such as question answering and knowledge graph construction.
Currently, prevailing NER methods are fully supervised models trained on a substantial amount of human-annotated data.
However, annotating sentences with named entity labels is difficult, expensive, and time-consuming \cite{Boecking.2021.iws, zhang-2021-wrench}.
The manually annotated dataset is also hard to expand---to insert new entity types or append new sentences, annotators must go over each sample, the effort of which is proportional to the data size.
\emph{Weak supervision} is proposed in recent works \citep{hoffmann-etal-2011-knowledge, Ratner-2017-Snorkel} to address this difficulty.
Instead of human annotations, weakly supervised methods generate multiple sets of \emph{weak annotations} using labeling functions (LFs, \aka weak supervision sources).
They can be knowledge bases, domain-specific dictionaries, pattern matching regular expressions or out-of-domain supervised models \cite{lison-etal-2020-named}.
Constructing LFs is significantly cheaper than annotating tokens manually.
And the weak annotations are expandable---the LFs can automatically annotate incoming sentences without any extra human labor.

\begin{figure*}[t!]
  \centering{
    \subfloat[The annotations and the corresponding true emission matrix of \emph{one} example LF]{%
      \label{fig:lf.examples}%
      \includegraphics[height=1.7in]{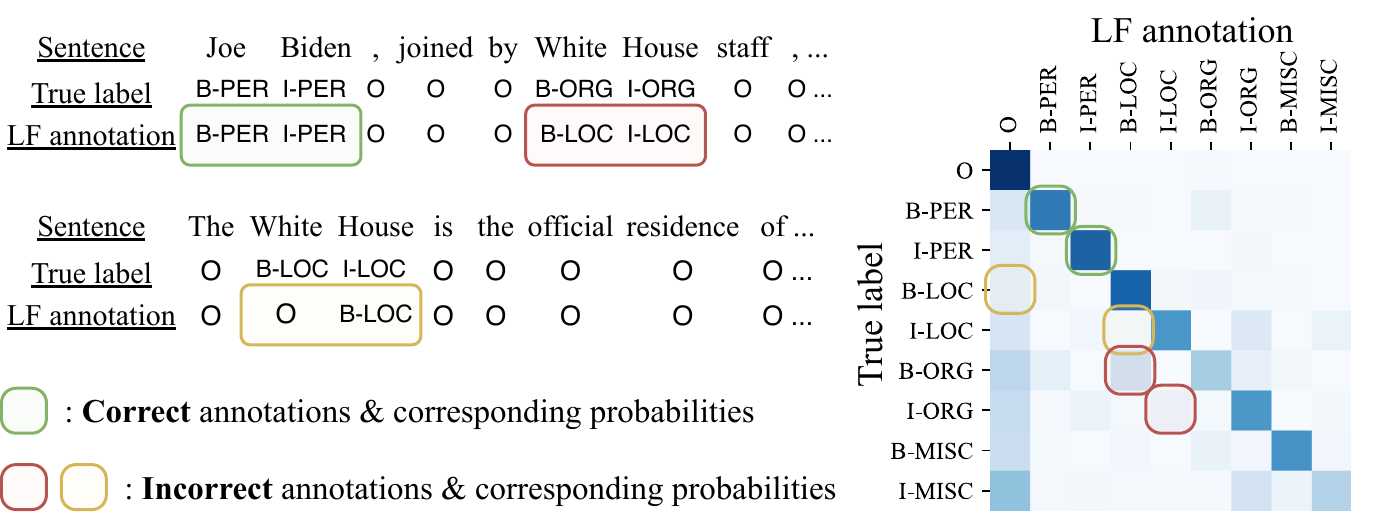}%
    }\hfil
    \subfloat[Model structures of CHMM and \ours]{%
      \label{fig:reg-chmm}%
      \includegraphics[height=1.7in]{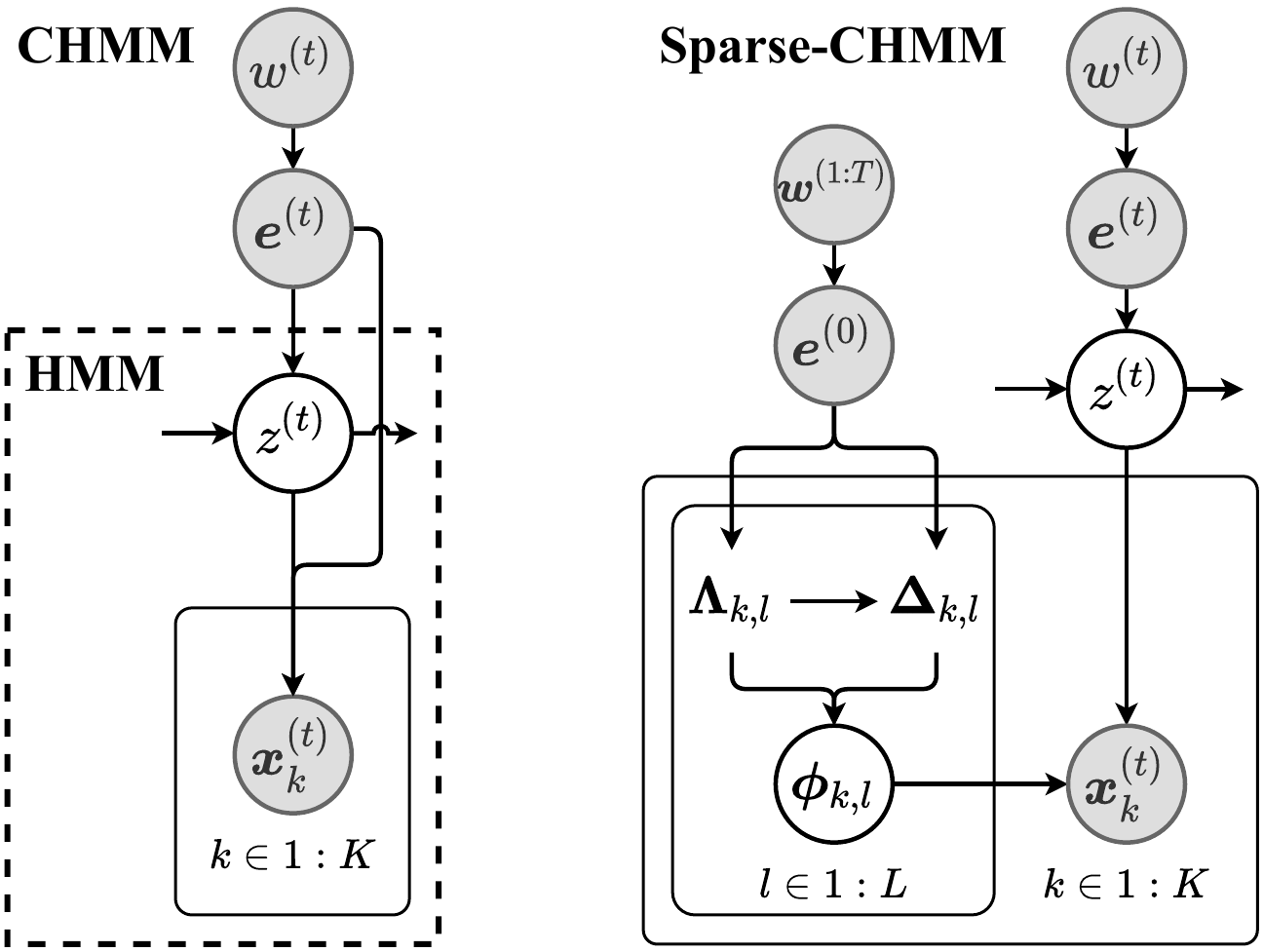}%
    }
  }
  \caption{
    \protect\subref{fig:lf.examples} shows an example of the weak annotations for weakly supervised NER and HMM's corresponding true emission pattern;
    \protect\subref{fig:reg-chmm} illustrates \ours's model structure.
    $\bLambda_k$ and $\bDelta_k$ in \protect\subref{fig:reg-chmm} respectively parameterize the diagonal and off-diagonal emission elements in \protect\subref{fig:lf.examples}, which are LF's probabilities of generating correct and incorrect labels.
  }
  \label{fig:example.and.model}
  \Description{Case study}
\end{figure*}

Nonetheless, it is challenging to design \emph{label models} which aggregate the LF annotations without seeing any ground-truth labels.
LFs are usually simple, and their annotations often suffer from incompleteness (low coverage/recall) and inaccuracy (low precision).
Moreover, the LFs can conflict with each other, generating contradictory results \cite{Ratner-2016-data-programming}.
Without true labels, solving the conflicts and reducing the annotation noise become a tricky task.

Some weakly supervised NER methods such as majority voting (MV) and Snorkel \citep{Ratner-2017-Snorkel} classify each token in the sentence independently.
Such approaches neglect the complex inter-word relationship of natural language.
Recent works \citep{nguyen-etal-2017-aggregating, Safranchik-etal-2020-weakly, lison-etal-2020-named, li-etal-2021-bertifying, parker-yu-2021-named} parameterize the label sequences with graphical models, regarding the unobserved ``true labels'' as latent variables and inferring them by optimizing the probabilities of the observed weak annotations through unsupervised learning.
\citet{nguyen-etal-2017-aggregating}, \citet{Safranchik-etal-2020-weakly} and \citet{lison-etal-2020-named} apply the hidden Markov models (HMMs) to capture the token dependency with the transition matrix.
However, the Markov property constrains HMMs' modeling of such dependency to only consecutive tokens meanwhile largely ignoring sentence semantics.
There are attempts to integrate neural networks (NNs) with graphical models \citep{li-etal-2021-bertifying, parker-yu-2021-named}.
\citet{li-etal-2021-bertifying} propose the conditional HMM (CHMM) that uses BERT token embeddings \cite{devlin-etal-2019-bert} to predict HMMs' transition and emission probabilities with NNs.
As each token embedding encodes the contextual semantics of the whole sentence, CHMM incorporates more comprehensive context information and alleviates the Markov constraint.
A similar approach is DWS \citep{parker-yu-2021-named}, which substitutes HMM by the conditional random field (CRF) and predicts the latent variables instead of their transitions.
The problem with both CHMM and DWS is that they directly predict all elements in the emission matrices from the embeddings.
When the number of LFs or labels is large, doing so requires a myriad of network parameters.
As the optimization is non-convex, this creates many local optima, making the model hard to train.
Besides, the training and inference of a large NN are much more computationally demanding, which hurts model efficiency.

In this work, we propose \ours.
Maintaining the transition scheme, \ours enhances CHMM by restricting its emission process with the empirical structure of an emission matrix (Figure~\ref{fig:lf.examples}).
Generally, the diagonal of the emission matrix is its most essential component, which represents the reliability of the LF annotations.
Considering this fact, \ours first focuses on estimating the LF reliability scores instead of the entire emission matrix.
Then it expands the sparse scores to the full-fledged emission matrix by placing the scores at the diagonals and filling up other vacancies with our designed expansion functions (\cref{subsec:basic-emiss}).
Compared with CHMM, \ours features sparser emission prediction process with much fewer parameters to learn, making the model easier to optimize and more efficient to train.

To enhance \ours's estimation of off-diagonal emission elements, we introduce another component---the weighted XOR (WXOR) scores (\cref{subsec:addon-emiss}).
The off-diagonal elements model LFs' probabilities of generating incorrect entity annotations.
These probabilities are typically small but sometimes comparable with or larger than the diagonal elements.
In such cases, using the diagonal-oriented reliability expansion functions alone is biased and inflexible.
To address this issue, we leverage LF annotations and the predicted LF reliabilities to calculate LFs' mutual disagreement frequencies as WXOR scores to refine off-diagonal emission elements.

Considering the amount of uncertainty in the emission, we treat the function-induced deterministic values as priors, with which a Dirichlet distribution is parameterized.
We then sample our final emission probabilities from the distribution (\cref{subsec:emiss-sampling}).
Previous works \citep{Kingma-2014-VAE, chem-2021-DrNAS} show that incorporating the probability model expands the search space, regularizes the gradient flow, and further prevents the model from being trapped by local optima.

In addition, to improve learning \ours's inter-correlated components, we introduce a three-stage training strategy.
Each stage optimizes only a subset of model parameters (\cref{subsec:training-procedure}) to reduce mutual interference and promote the training stability.

In summary, our contributions include:
\begin{itemize}[leftmargin=*]
  \item proposing \ours to estimate LF reliability scores from sentence embeddings, and expand the scores to sparsely predicted emission matrices with our designed expansion functions;
  \item modeling the probabilities of an LF generating incorrect annotations with the WXOR scores to augment the estimation of emission off-diagonal elements;
  \item a three-stage training strategy and the Dirichlet sampling process to improve training stability and model robustness; and
  \item thorough experiments on $5$ comprehensive datasets with SOTA baselines that demonstrate the superiority of \ours, and ablation studies that justify the design of model components.
\end{itemize}

The code and data are available at \href{https://github.com/Yinghao-Li/Sparse-CHMM}{github.com/Yinghao-Li/Sparse-CHMM} for reproducibility.

\section{Problem Definition}
\label{sec:definition}

In this section, we define the weakly supervised NER task.
Suppose we have $T$ tokens $\w^{(1:T)}$ in the input sentence and a set of candidate entities $\ents$ with size $E$, NER models assign one entity label to each token $\w^{(t)}, t\in 1:T$.
Using the BIO tagging scheme, the label set is $\lbs = \{ \texttt{O} \} \cup \{ \texttt{B-}\ent, \texttt{I-}\ent\}_{\ent\in\ents}$ with the size of $L=2E+1$.
Label ``\lbfont{O}'' indicates that the current token belongs to no pre-defined entities, which is always indexed by $1$ in the following discussion.
An example of NER labels is shown in Figure~\ref{fig:lf.examples}.

For weakly supervised NER, we have $K$ independent LFs, each providing a sequence of weak annotations $\x^{(1:T)}_k$.
$\x^{(t)}_k \in \{0,1\}^{L}$ is one-hot over the label set $\lbs$.
Label models aim to approximate the ground-truth labels $\y^{(1:T)} \in \lbs^T$ with latent states $\z^{(1:T)} \in \lbs^T$ given the input tokens and weak annotations $\{\w^{(1:T)}, \x^{(1:T)}_{1:K} \}$.

In the following discussion, we do not distinguish the label string and label index, and uniformly represent them by integers $l$, $i$, or $j\in 1:L$.
In addition, we focus on one sentence and omit the sentence index $m \in 1:M$ unless specified otherwise.

\begin{figure*}[t!]
  \centerline{\includegraphics[width = 0.98\textwidth]{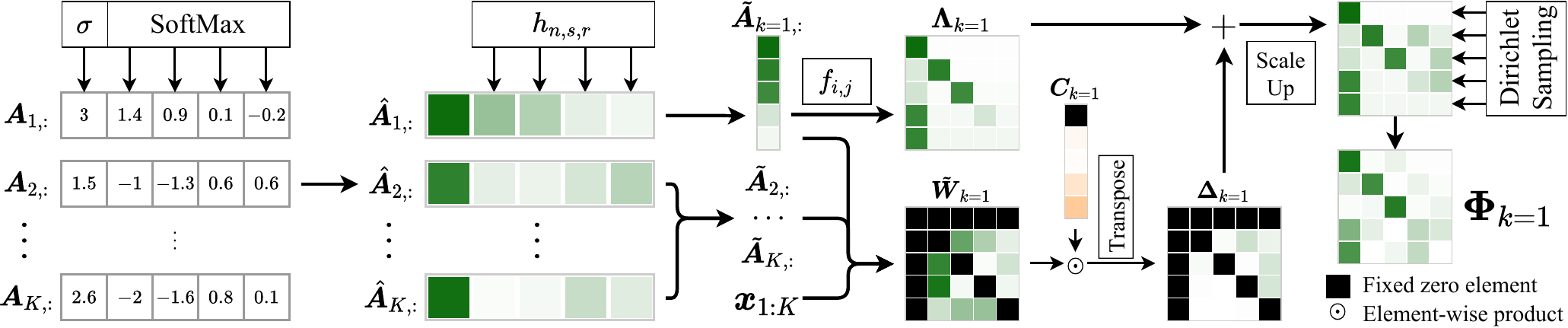}}
  \caption{
    Emission construction pipeline for LF $k=1$ with $L=5$ labels.
    Darker colors indicate larger values in range $[0,1]$.
  }
  \Description{Emission Matrix}
  \label{fig:emiss.constr}
\end{figure*}

\section{Method}
\label{sec:method}

\ours has three main components:
1) the transition matrix $\bPsi$ (\cref{subsec:transition}),
2) the \emph{emission base prior} matrix $\bLambda$ (\cref{subsec:basic-emiss}), and
3) the \emph{emission addon prior} matrix $\bDelta$ (\cref{subsec:addon-emiss}).
The transition matrix controls the probability of moving from one latent state to another.
$\bLambda$ and $\bDelta$ are the components of the emission matrix $\bPhi$ (\cref{subsec:emiss-sampling}), each element in which $\Phi_{k,i,j} \triangleq p(x_{k,j}^{(t)}=1|z^{(t)}=i)$ defines the probability of LF $k$ observing label $j$ when the true label is $i$.
$\bLambda$ focuses on the emission diagonals, whereas $\bDelta$ refines the off-diagonal values.
In \cref{subsec:training}, we describe the training strategy and present a three-stage pipeline that increases training stability and maximizes the model performance.

\subsection{Transition}
\label{subsec:transition}

Same as \citet{li-etal-2021-bertifying}, we directly predict the \emph{token-wise} transition probabilities $\Psi^{(t)}_{i,j} \triangleq p(z^{(t)}=j|z^{(t-1)}=i, \e^{(t)})$ from the BERT \emph{token} embeddings $\e^{(t)}$.
Specifically, we input the embeddings into one fully connected (FC) layer of NN and reshape the output vectors into $L\times L$ matrices.
A SoftMax function is applied along the columns to make each row a point in an $L$-dimensional probability simplex.
\begin{gather*}
  \bm{S}^{(t)} \in \R^{L\times L} = \text{reshape}(\fc(\e^{(t)})); \\
  \bPsi_{l,:}^{(t)} \in (0,1)^L = \softmax(\bm{S}_{l,:}^{(t)}).
\end{gather*}
Here $\bm{S}^{(t)}$ is an intermediate variable, and $l$ is the row label index.

\subsection{Emission}
\label{subsec:emission}

Instead of predicting every element in the emission matrices $\bPhi^{(t)} \in [0,1]^{K \times L \times L}$, \ours predicts LF reliabilities $\tilde{\A} \in [0,1]^{K \times L}$ with NN and expands it to the base prior $\bLambda \in [0,1]^{K \times L \times L}$ through pre-defined functions (\cref{subsec:basic-emiss}).
From $\tilde{\A}$ and LF observations $\x$, we also calculate the WXOR scores $\tilde{\W} \in [0,1]^{K \times L \times L}$ that track the probabilities of LFs' giving incorrect annotations.
$\tilde{\W}$ is then scaled by another NN predicted matrix $\C\in[0,1]^{K\times L}$ to form the addon prior $\bDelta$ (\cref{subsec:addon-emiss}).
We sample the emission matrix $\bPhi$ from a Dirichlet distribution parameterized by $\bLambda$ and $\bDelta$ to facilitate model training (\cref{subsec:emiss-sampling}).
The pipeline is illustrated in Figure~\ref{fig:emiss.constr}.

Compared with the transition, the emission variance is much smaller \wrt the input tokens.
So we predict the \emph{sentence-level} emission-related variables from the BERT \emph{sentence} embedding $\e^{(0)}$ \citep{devlin-etal-2019-bert}, which deprecates their token indicator ``$\cdot^{(t)}$''.

\begin{figure}[tbp]
  \centerline{\includegraphics[width = 0.45\textwidth]{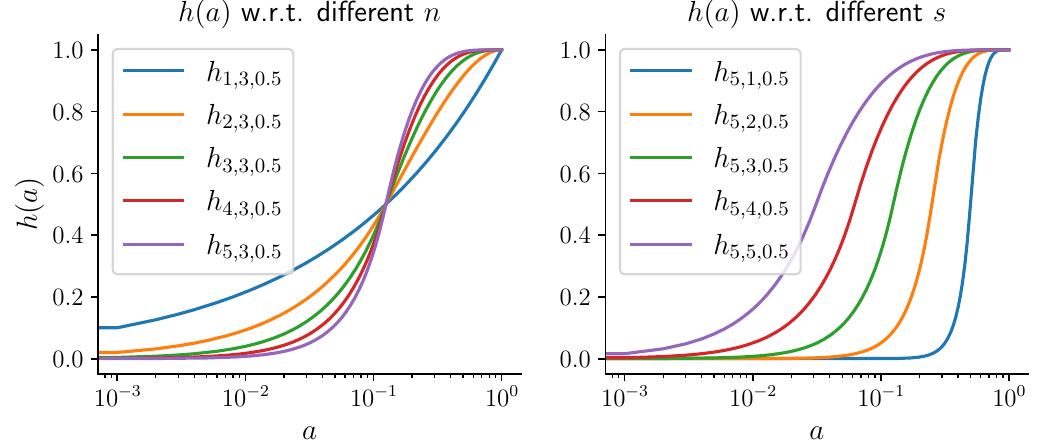}}
  \caption{
    Reliability scaling function $h_{n,s,r}(a)$ \wrt its exponential terms $n$ and $s$.
    $s>1$ scales up all values, whereas $n>1$ suppresses small values but augments large ones.
  }
  \label{fig:ha}
  \Description{Reliability scaling function.}
\end{figure}

\subsubsection{Labeling Function Reliabilities and Emission Base Prior}
\label{subsec:basic-emiss}

For LF $k$, the diagonal elements of the emission matrix 
\begin{equation*}
  \Phi_{k, l, l} \triangleq p(x_{k,l}^{(t)}=1|z^{(t)}=l, \e^{(0)}), \quad l \in 1:L,
\end{equation*}
as shown in Figure~\ref{fig:lf.examples}, has a distinctive physical meaning: the probabilities of LF $k$ observing the correct labels, \ie, the reliabilities of LF $k$.
In other words, we can estimate the performance of LF $k$ given its emission matrix $\bPhi_k$, or vice versa: \emph{we can construct a plausible emission matrix given the knowledge of how an LF performs.}

With this concept established, we start by predicting the reliability logits $\A \in \R^{K\times L}$ from the sentence embedding:
\begin{equation*}
  \A = \text{reshape}(\fc(\e^{(0)})),
\end{equation*}
which is to be placed at the diagonal of $\bPhi$ after normalization.\footnote{
  On datasets with more than one entity types, we adopt entity-level reliability scores, \ie, $A_{k,\lbfont{B-}\ent}=A_{k,\lbfont{I-}\ent},$.
  This reduces the model size and training difficulty.
}
However, as LFs observe more label \texttt{O} (indexed by $1$) than other labels, the model tend to emphasize the corresponding weight $\Phi_{k,l,1} \triangleq p(x_{k,1}^{(t)}=1|z^{(t)}=l, \e^{(0)})$ in the emission.
Because the vector $\bPhi_{k,l}$ is from a simplex and sums up to $1$, large $\Phi_{k,l,1}$ results in unreasonably small diagonal values $\Phi_{k,l,l}$.
Fortunately, this can be fixed by applying SoftMax along the LFs:
\begin{equation}
  \label{eq:softmax-A}
  \begin{gathered}
    \hat{\A}_{:, l} = \softmax(\A_{:, l}), \quad l\geq 2.
  \end{gathered}
\end{equation}
It guarantees that at least one LF has a high score, preventing the model from neglecting all weak annotations simultaneously.
Since ${\A}_{k, 1}$ is the emission confidence to \lbfont{O} itself, we only need to constrain its value to $(0,1)$ through the element-wise sigmoid function $\sigma$:
\begin{equation*}
  \label{eq:sigmoid-A}
  \hat{\A}_{:, 1} = \sigma( \A_{:, 1} ).
\end{equation*}

However, SoftMax in \eqref{eq:softmax-A} enforces $\sum_k \hat{\A}_{k, l} = 1$.
This correlation establishes a \emph{fixed-size} pool from which reliability scores are drawn, which harms the model flexibility as it fails the cases where multiple LFs are confident, or no LF is confident.
To de-correlate the scores, we introduce a piecewise scaling function (illustrated in Figure~\ref{fig:ha}):
\begin{equation}
  \label{eq:hnsr}
  \begin{gathered}
    \tilde{A}_{k, l} = h_{n,s,r}(\hat{A}_{k, l}); \\
    h_{n,s,r}(a)= 
    \begin{cases}
      \frac{1}{r^{n-1}} a & a^\frac{1}{s} < r; \\
      -\frac{1}{(1-r)^{(n-1)}}(1-a^\frac{1}{s})^n + 1 &  a^\frac{1}{s} \geq r.
    \end{cases}
  \end{gathered}
\end{equation}
It calibrates the SoftMax output with scales defined by exponents $n$ and $s$.
To enable more subtle control of the scores, we utilize the split point $r \in [0, 1]$ to segment the input domain into upper and lower halves, each scaled differently.\footnote{\label{criteria}Please refer to \cref{appsec:function.criteria} for design principles.}

\begin{figure}[tbp]
  \centerline{\includegraphics[width = 0.45\textwidth]{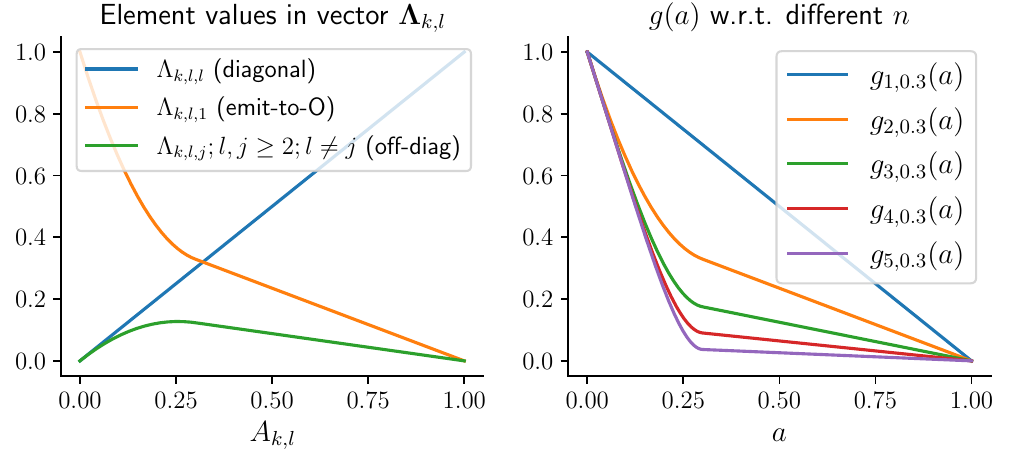}}
  \caption{
    Left: the elements in $\bLambda_{k,l}, \forall k, l$ \wrt the reliability score $A_{k,l}$ with $L=5, r=0.3, n=2$.
    Right: $g_{n,r}(a)$ \wrt exponent $n$.
    Larger $n$ decreases emit-to-\lbfont{O} probabilities faster.
  }
  \Description{reliability expansion function}
  \label{fig:ga}
\end{figure}

We then expand $\tilde{\A}$ to the \emph{emission base prior} matrix $\bLambda \in [0,1]^{K \times L \times L}$ through expansion functions defined according to the latent state $z^{(t)} = i$ and the observation $x_{k,j}^{(t)} = 1$:
\begin{equation*}
  \Lambda_{k,i,j} = f_{i,j}(\tilde{A}_{k,i});\quad \forall i,j \in 1:L,
\end{equation*}
\begin{equation}
  \label{eq:fij}
  f_{i,j}(a) = 
  \begin{cases}
    a & i=j; \\
    \frac{1}{L-1}(1-a) & i=1,j\geq 2; \\
    g_{n,r}( a ) & i \geq 2, j=1; \\
    \frac{1}{L-2}(1-a-g_{n,r}( a )) & i \geq 2, j \geq 2, i\neq j,
  \end{cases}
\end{equation}
\begin{equation}
  \label{eq:gnr}
  g_{n,r}( a ) =
  \begin{cases}
    \frac{2-L}{(n-1)r^n-nr^{n-1}}a^n + (1-L)x + 1 & a \leq r; \\
    \frac{g_{n,r}( r )}{r-1}a - \frac{g_{n,r}( r )}{r-1} & a > r,
  \end{cases}
\end{equation}
where $n$ is the exponential term and $r$ is the split point.
The reliability scores $\tilde{\A}_k$ of LF $k$ are placed at the diagonal of $\bLambda_k$ (\ie, function $f_{i=j}$).
When the latent state is \lbfont{O} ($z^{(t)}=i=1$), the probabilities of emitting to non-\lbfont{O} $p(\x^{(t)}_{k,j} = 1 | z^{(t)}=1, \e^{(0)}),\ \forall j\geq 2$ are uniform and sum up to $1-\tilde{A}_{k,1}$.
When the latent state is not \lbfont{O} ($z^{(t)}=i \geq 2$), an LF statistically inclines toward observing \lbfont{O} than other entities,
making emit-to-\lbfont{O} probabilities $p(\x^{(t)}_{k,1} = 1 | z^{(t)}=i, \e^{(0)}),\ \forall i\geq 2$ larger and more important than other off-diagonal values $p(\x^{(t)}_{k,j} = 1 | z^{(t)}=i, \e^{(0)}),\ \forall i,j\geq 2, i\neq j$, as shown in Figure~\ref{fig:lf.examples}.
Therefore, we specify $g_{n,r}$ to emphasize emit-to-\lbfont{O} probabilities.\footnoteref{criteria}
With small reliability scores, an LF is less confident about choosing which label, so the Uniform off-diagonal non-\lbfont{O} values are set close to the diagonals.
When the scores are above threshold $r$, the LF is considered confident enough to abate the emission probabilities to other entities.
The exponent $n$ controls how fast the emit-to-\lbfont{O} probabilities decrease, \ie, how close off-diagonal values to the diagonal before $r$.
The functions $f$ and $g$ are illustrated in Figure~\ref{fig:ga}.

\subsubsection{Weighted XOR and Emission Addon Prior}
\label{subsec:addon-emiss}

In contrast to the diagonals $\Phi_{k,l,l}$, the off-diagonal values $\Phi_{k,i,j}; i,j\geq 2; i\neq j$ are the probabilities of LF $k$ \emph{misclassifying} label $i$ as $j$.
Such probabilities are generally insignificant.
But in some cases, they can be prominent (as the example shown in Figure~\ref{fig:lf.examples}) and affects the model performance.
The base prior $\bLambda$ assigns equal weights to all off-diagonals and fails to adjust to such cases.

Therefore, we develop the WXOR scores $\tilde{\W} \in [0,1]^{K \times L \times L}$ to capture the misclassification probabilities.
Given the annotations of LF $k$ and other LFs $k'\in 1:K\backslash k$, a WXOR logit is specified between the query label $\lquery \in 2:L$ and the target label $\ltgt \in 2:L$:
\begin{equation}
  \label{eq:wxor}
  \begin{gathered}
    W^{(t)}_{k, \lquery, \ltgt} = (1-\tilde{A}_{k, \lquery})x^{(t)}_{k, \lquery} \sum_{k'=1}^K \tilde{A}_{k', \ltgt}x^{(t)}_{k',\ltgt}; \\
    \W^{(t)}_{k, 1, :} = \W^{(t)}_{k, :, 1} = \0; \quad W^{(t)}_{k, l, l} = 0,\ \forall l\in 1:L.
  \end{gathered}
\end{equation}
It depicts the likelihood that LF $k$ makes mistakes in its observation of $\lquery$ \wrt the other LFs at time-step $t$.
$1-\tilde{A}_{k, \lquery}$ is the \emph{uncertainty level} of LF $k$ to its observation of $\lquery$;
$\tilde{A}_{k', \ltgt}$ is the confidence of LF $k'$ to $\ltgt$;
and the binary $x^{(t)}_{k, l}$ decides whether $l$ is observed.
$W^{(t)}_{k, \lquery, \ltgt}$ is large when LF $k$ observes $\lquery$ but is uncertain, while other LFs are confident about observing $\ltgt$.
As $W^{(t)}_{k, \lquery, \ltgt}$ is non-zero \iff LF $k$ does but other LFs do not observe $\lquery$, it is similar to the XOR gate and hence gets its name.
Label \lbfont{O} and the diagonal need no WXOR as these elements are well-defined in $\bLambda$, so the corresponding values in $\W$ are fixed to $0$.

$\W^{(t)}$ tensors are then summed up and normalized to form the ubiquitous WXOR:
for $M$ sentences with length $T_m,m\in 1:M$ each,
\begin{equation*}
  \hat{W}_{k, \lquery, \ltgt} = \frac{\sum_{m=1}^M \sum_{t=1}^{T_m} {W}^{(t)}_{m, k, \lquery, \ltgt}}{\sum_{m=1}^M \sum_{t=1}^{T_m} {x}^{(t)}_{m, k, \lquery}}.
\end{equation*}

Combining $\hat{\W}$ with the emission base prior $\bLambda$ is also nontrivial.
Straightforward summation is unpractical since the relative scale of $\hat{\W}$ and $\bLambda$ is obscure due to the summation in \eqref{eq:wxor}.
In regard of this, we leverage another scaling factor $\C \in (0,1)^{K \times L}$ predicted from the BERT sentence embedding through one layer of NN:
\begin{equation*}
  \C = \text{reshape}(\sigma(\fc(\e^{(0)}))).
\end{equation*}
Its domain $(0,1)$, however, enables $\C$ to down-scale values but not up-scale them.
Consequently, we need to properly adjust the elements of $\hat{\W}$ so that they have large enough pre-scaling values:
\begin{equation*}
  \tilde{\W}_{k, :, \ltgt} = \softmax (\hat{\W}_{k, :, \ltgt}).
\end{equation*}

With the established matrices, we construct the \emph{emission addon prior} $\bDelta \in [0,1]^{K\times L \times L}$ by re-scaling the rows of $\tilde{\W}_{k}$:
\begin{equation}
  \bDelta_{k,:,l} = C_{k,l} \times \tilde{\W}_{k,l,:}.
\end{equation}
Notice that in $\bDelta_{k}$, the target labels are at the first axis and the query labels second, different from $\tilde{\W}_k$.
As the target labels are essentially the latent states we want to predict, this unifies the physical meaning of $\tilde{\W}$ and the emission matrix $\bPhi$.

\subsubsection{Emission Matrix Sampling}
\label{subsec:emiss-sampling}

Instead of using the deterministic priors $\bLambda$ and $\bDelta$ directly, we choose to sample the latent emission variables from the Dirichlet distribution parameterized by the priors.
This is a common practice in graphical models considering that proper latent distributions are more representative and robust to noise than the deterministic priors \citep{Kingma-2014-VAE, chem-2021-DrNAS}.
The random sampling helps the model escape the saddle points or local optima.
In addition, Dirichlet distribution samples from probability simplex without requiring its concentration parameters to sum up to one, which makes the parameters selection more flexible.

To construct the concentration parameters $\bOmega \in \R_+^{K\times L \times L}$, we add the base prior and the addon prior together and scale the results:
\begin{equation*}
  \bOmega = \nu^{\rm expan} \times (\bLambda + \bDelta) + \nu^{\rm base}.
\end{equation*}
$\nu^{\rm base}\in \R_+$ and $\nu^{\rm expan}\in \R_+$ controls the minimum concentration value and the concentration range, respectively.
Larger $\nu_{\rm expan}$ results in smaller sampling variance.
Each row of the emission matrix $\bPhi_{k}$ of LF $k$ is sampled independently from the distribution:
\begin{equation*}
  \bPhi_{k,l} \sim \Dir (\bOmega_{k,l} ).
\end{equation*}
We use pathwise derivative estimators developed by \citet{Jankowiak-2018-pathwise-derivatives} to push the gradient back through the Dirichlet.

Notice that we only apply Dirichlet sampling when it requires backpropagation.
On other occasions, such as validation and test, the samples are substituted by the mean of the Dirichlet distribution.

\subsection{Model Training}
\label{subsec:training}

\ours is an unsupervised model whose optimization does not leverage any ground-truth labels $\y$ but only the weak labels $\x$.
Section~\ref{subsec:initialization}--\ref{subsec:training-procedure} describe \ours's training strategy, whereas \cref{subsec:training.complexity} analyzes the inference complexity.

\subsubsection{Model Pre-training}
\label{subsec:initialization}

A good initialization is critical for HMMs to gain good performance.
However, as \ours contains several NNs, assigning proper initial values to model parameters is unrealistic.
Thus, we adopt the approach used by \citet{li-etal-2021-bertifying}, which pre-trains the neural networks by minimizing the Euclidean distances between the predicted transitions and emissions and the initial matrices $\bPsi^*$ and $\bPhi^*$ that are from the observation statistics.
The objective is the mean squared error (MSE) loss:
\begin{equation}
  \label{eq:model.initialization}
  \ell_{\rm MSE} = \frac{1}{K} \sum_{k=1}^K \|\bPhi_k - \bPhi^*\|_F^2 + \frac{1}{T}\sum_{t=1}^T \|\bPsi^{(t)} - \bPsi^*\|_F^2,
\end{equation}
where $\|\cdot\|_F$ is the Frobenius norm.
Different from \citep{li-etal-2021-bertifying} or \citep{lison-etal-2020-named}, we require no prior knowledge of LF performance.

\subsubsection{Training Objective}
\label{subsec:objective}

Same as CHMM \citep{li-etal-2021-bertifying}, \ours is trained using the generalized EM algorithm.
Specifically, we leverage the gradient ascent method in the M step to optimize the expected complete data log likelihood computed in the E step:
\begin{equation}
    \label{eq:em.q}
    Q(\btht, \btht^{\rm old}) \triangleq \expt_{\z} [\ell_c (\btht) | \x, \btht^{\rm old}].
\end{equation}
The expectation is defined over the hidden states given LF observations and the current model parameters $p(\z^{(1:T)} | \x^{(1:T)}, \btht^{\rm old})$.
The data log likelihood $\ell_c({\btht})$ is defined as:
\begin{equation}
  \label{eq:lc}
  \begin{aligned}
    & \ell_c(\btht) \triangleq \log p(\z^{(0:T)}, \x^{(1:T)} | \btht, \e^{(0:T)}) = \log p(z^{(0)}) +\\
    &\quad \sum_{t=1}^T \log p(z^{(t)}|z^{(t-1)}, \e^{(t)}) + \sum_{t=1}^T \log p(\x^{(t)}|z^{(t)}, \e^{(0)}),
  \end{aligned}
\end{equation}
of which the conditional independency comes from the Markov assumption.
The computation of \eqref{eq:em.q} largely aligns with CHMM.
One difference is the dependency of the emission evidence:
\begin{equation}
  \label{eq:varphi}
  \varphi_l^{(t)} \triangleq p(\x^{(t)}|z^{(t)}=l, \e^{(0)})=\prod_{k=1}^K \sum_{j=1}^L \Phi_{k,l,j}x_{k,j},
\end{equation}
which in our case is sentence-level $\e^{(0)}$, whereas in CHMM is token-level $\e^{(t)}$.
This essentially is attributed to the different approaches to obtain the emission matrix $\bPhi$ (\cref{subsec:emission}).
The computation of \eqref{eq:em.q} and other forward inference details are in \cref{appsec:objective}.

\begin{figure}[t!]
  \centerline{\includegraphics[width = 0.4\textwidth]{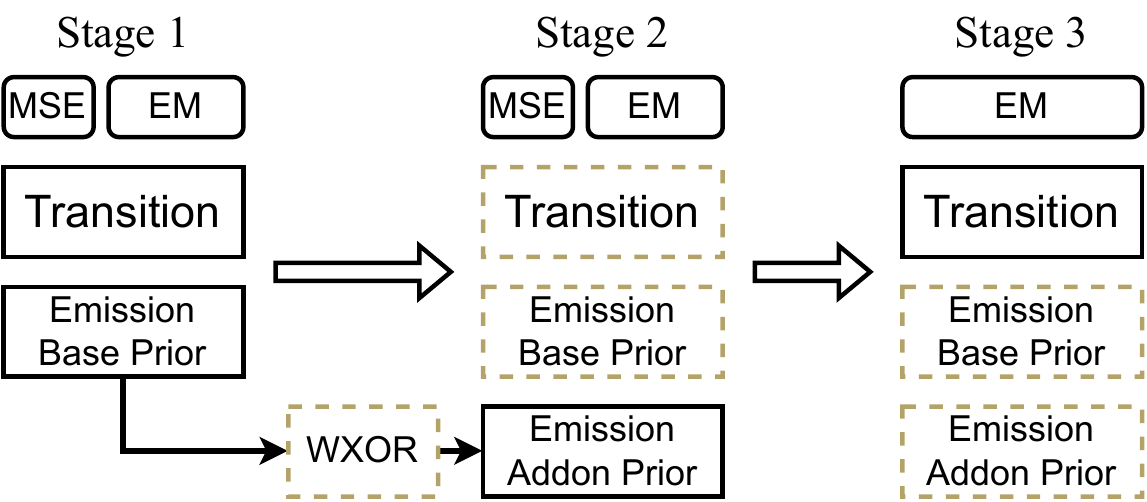}}
  \caption{
    \ours training procedure.
    The model parameters of the dotted squares are frozen.
    MSE and EM are optimization approaches associated with the model pre-training and training steps.
  }
  \label{fig:training-process}
  \Description{Training process.}
\end{figure}

\subsubsection{Training Procedure}
\label{subsec:training-procedure}

As the computation of the emission is complicated and inter-dependent, we adopt a three-stage training strategy to allow sufficient training of each model component, as shown in Figure~\ref{fig:training-process}. 
It decouples the optimization of the model components while keeping the training efficient.

Stage 1 is focused on optimizing the transition matrix $\bPsi$ and the emission base prior $\bLambda$, excluding the addon prior by setting $\bDelta = \0$.
The reason is straightforward:
calculating the WXOR scores $\tilde{\W}$ requires high-quality reliability scores $\tilde{\A}$, which are obtained by optimizing $\bLambda$ (\cref{subsec:training}).
Stage 2 trains the addon prior $\bDelta$, leaving $\bPsi$ and $\bLambda$ frozen.
This stage aims to search for the best scaling factors $\C$ for $\tilde{\W}$, and freezing the irrelevant parameters relieves the training pressure.
$\tilde{\W}$ is calculated right after stage 1 with all training and validation instances and remains constant for stages 2 and 3.

Stage 3 is inspired by our empirical discovery about HMMs.
We find that if we only optimize the transition matrix $\bPsi$ with the emission fixed, generally to the true values, the model performance can be improved.
The true emission is the statistics of the annotations $\x_k$ of LF $k$ given ground-truth labels $\y$:
\begin{equation*}
  \Phi^{\rm true}_{k,i,j} = p(\x_k|\y) \triangleq \frac{\sum_{m=1}^M \sum_{t=1}^{T_m} \I(x^{(t)}_{m,k,j}=1, y^{(t)}_{m}=i) }{\sum_{m=1}^M \sum_{t=1}^{T_m} \I(y^{(t)}_m=i) },
\end{equation*}
where $\I(\cdot)$ is the indicator function.
Consequently, we believe that continuing training the transition of \ours for several more epochs with the emission frozen would lead to a similar performance improvement.

In addition, stages 1, 2, and 3 use different pre-training strategies.
The pre-training of stage 1 is the same as \cref{subsec:initialization}.
Stage 2's pre-training is to initialize the NN parameters associated with the addon prior $\bDelta$ only, so we drop the transition part of the MSE loss and substitute the target emission by
$\bPhi' = \lambda \bPhi^* + \frac{(1-\lambda)}{M}\sum_{m=1}^M \bPhi_m^{(\rom{1})}$
to take advantage of stage 1's results.
Here $\lambda \in [0,1]$ is the weight of observation statistics, which is fixed to $0.2$ in our experiments.
$\bPhi_m^{(\rom{1})}$ is the optimized emission from stage 1.
Stage 3 successes all model parameters from the previous stage and thus has no pre-training.


\subsubsection{Complexity Analysis}
\label{subsec:training.complexity}

Compared with CHMM, \ours significantly reduces the training resource consumption by predicting sparse emission elements.
The emission NN parameter number and computation complexity are shown in Table~\ref{tb:inference.complexity}, where the transition attributes are not presented because they are the same for both models.
The factor $2$ for \ours comes from matrices $\tilde{\A}$ and $\C$.
We can see that \ours reduces the emission NN parameter number to $2/L$ of CHMM and the complexity to $1/(T\times L)$, which are substantial when the number of entity labels $L$ is large.
The complexity of other emission elements, \ie, \eqref{eq:hnsr}--\eqref{eq:wxor}, is negligible because they do not contain the embedding dimension $d^{\rm emb}$, which is much larger than other terms.
Calculating the WXOR scores can be as complex as $\O(M\times T\times K^2 \times L^2)$, but they are calculated only once at stage 2 and stay fixed henceforth.

\begin{table}[t!]\small
  \centering
  \caption{
    \label{tb:inference.complexity}
    Emission complexity of each epoch.
  }
  \begin{threeparttable}
    \begin{tabular}{c|c|c}
      \toprule
       & CHMM & \ours \\
      \midrule
      \# NN PRM & $d^{\rm emb} \times K \times L^2$ & $2 \times d^{\rm emb} \times K \times L$ \\
      NN CPL & $\O(M \times T \times d^{\rm emb} \times K \times L^2)$ & $\O(M \times d^{\rm emb} \times K \times L)$ \\
      \bottomrule
    \end{tabular}
    \begin{tablenotes}
      \footnotesize{
        \item PRM is short for ``parameters''; CPL is short for ``complexity''.
        \item $M$ is the number of training sentences, $T$ is the average of sentence lengths, and $d^{\rm emb}$ is the dimension of the BERT embeddings.
      }
    \end{tablenotes}
  \end{threeparttable}
\end{table}

\section{Experiments}
\label{sec:experiment}


\subsection{Experiment Setup}
\label{subsec:exp.setup}

\subsubsection{Datasets}
We consider \num{5} NER datasets from the generic domains to the biomedical and chemical domains:
1) \textbf{CoNLL 2003} (English subset) \citep{tjong-kim-sang-de-meulder-2003-introduction} labels \num{22137} sentences from the Reuters news stories with \num{4} entity types, including \lbfont{PER}, \lbfont{LOC}, \lbfont{ORG} and \lbfont{MISC}.
2) The \textbf{LaptopReview} dataset \citep{pontiki-etal-2014-semeval} consists of \num{3845} sentences of laptop reviews.
The laptop-related \lbfont{Terms} are regarded as named entities.
3) The \textbf{NCBI-Disease} dataset \citep{Dogan-2014-NCBI} contains \num{793} PubMed abstracts annotated with \lbfont{Disease} mentions.
4) \textbf{BC5CDR} \citep{Li-2016-BC5CDR} consists of \num{1500} PubMed articles with \lbfont{Chemical} and \lbfont{Disease} entities.
5) \textbf{OntoNotes 5.0} \citep{Weischedel-2013-ontonotes} is a very large dataset containing \num{143709} sentences labeled with \num{18} fine-grained entity types.

We use the LFs included in the Wrench benchmark platform \citep{zhang-2021-wrench} for all datasets.
Table~\ref{tb:dataset.statistics} shows the dataset statistics, including the number of entities and labeling functions.
The performance of the LFs on the test dataset is listed in \cref{appsec:lf.metrics}.

\subsubsection{Baselines}
We compare our model with the following Wrench baselines:
1)~\textbf{Majority Voting} (MV) returns the label that has been observed by most LFs and chooses randomly from the tie if it exists.
2)~\textbf{Snorkel} \citep{Ratner-2017-Snorkel} is a context-free token-based simple graphical model which assumes the tokens are independent.
3)~\textbf{HMM}, used in \citep{nguyen-etal-2017-aggregating, lison-etal-2020-named, Safranchik-etal-2020-weakly}, is a popular weakly-supervised NER label model with certain context representation ability.
4)~\textbf{Conditional HMM} (CHMM, \citealp{li-etal-2021-bertifying}) augments HMM by predicting the token-wise transition and emission probabilities from the BERT token embeddings through NNs.
5)~\textbf{ConNet} \citep{lan-etal-2020-learning} uses the context-aware attention mechanism to aggregate the CRF representations of different LFs.\footnote{DWS \citep{parker-yu-2021-named} is similar to CHMM but has no open-source implementation.}

We also include \num{3} supervised methods as references:
1)~a fully supervised \textbf{BERT-NER} model trained with human annotations,
2)~the \textbf{best consensus} of LFs, which is an oracle that always selects the correct token annotations from the LFs; and
3)~\textbf{CHMM-FE}, which is CHMM with the fixed ground-truth emissions as described in \cref{subsec:training-procedure}.
Every supervised method accesses the true labels in one way or another during training.

\subsubsection{Evaluation Metrics}

The NER label models are evaluated using the \emph{micro}-averaged \emph{entity-level} precision, recall, and \fone scores.
We calculate the metrics with the \textbf{seqeval} Python package.\footnote{\href{https://github.com/chakki-works/seqeval}{https://github.com/chakki-works/seqeval}}
The results come from the average of \num{5} random trials.

\subsubsection{Implementation Details}

Following \citet{li-etal-2021-bertifying}, we apply different BERT variants to different datasets.
Specifically, we use bert-base-uncased \citep{devlin-etal-2019-bert} on CoNLL 2003, LaptopReview and OntoNotes 5.0, bioBERT \citep{Lee-2019-biobert} on NCBI-Disease and SciBERT \citep{beltagy-etal-2019-scibert} on BC5CDR.

In alignment with \citet{zhang-2021-wrench}, we evaluate \ours inductively.
The model is trained on the training dataset and tested on the test dataset.
The validation set is for early stopping and hyper-parameter fine-tuning.
In addition, the WXOR scores are calculated on the combination of training and validation datasets.
Please refer to \cref{appsec:parameters} for more implementation details and the model hyper-parameters that lead to the presented results.

\begin{table}[tbp]\small
    \caption{
    Dataset statistics.
    }
    \centering
    \begin{tabular}{c|c|c|c|c|c}
    \toprule
     & CoNLL & NCBI & BC5CDR & Laptop & OntoNotes \\
    \midrule
    \# Instance & \num{22137} & \num{793} & \num{1500} & \num{3845} & \num{143709} \\
    \# Training & \num{14041} & \num{593} & \num{500} & \num{2436} & \num{115812} \\
    \# Validation & \num{3250} & \num{100} & \num{500} & \num{609} & \num{5000} \\
    \# Test & \num{3453} & \num{100} & \num{500} & \num{800} & \num{22897} \\
    \midrule
    \# Entities & \num{4} & \num{1} & \num{2} & \num{1} & \num{18} \\
    \# LFs & \num{16} & \num{5} & \num{9} & \num{3} & \num{17} \\
    \bottomrule
    \end{tabular}
    \label{tb:dataset.statistics}
\end{table}

\begin{table*}[t!]\small
    \caption{
      Evaluation results on the test datasets, presented as ``\fone (Precision / Recall)'' in \%.
    }
    \centering
    \begin{threeparttable}
      \begin{tabular}{c|c|c|c|c|c|c}
        \toprule
        \multicolumn{2}{c|}{Models} & CoNLL 2003 & NCBI-Disease & BC5CDR & LaptopReview & OntoNotes 5.0 \\
        \midrule
        \multirow{3}{*}{\shortstack{Supervised\\Methods}}
        & BERT-NER & 90.74 (90.37 / 91.10) & 88.89 (87.05 / 90.82) & 88.81 (87.12 / 90.57) & 81.34 (82.02 / 80.67) & 84.11 (83.11 / 85.14) \\
        & Best consensus & 86.73 (98.62 / 77.39) & 81.65 (99.85 / 69.06) & 88.42 (99.86 / 79.33) & 77.60 (100.0 / 63.40) & 85.11 (97.35 / 75.61)\\
        & CHMM-FE & 71.43 (72.89 / 70.02) & 81.86 (90.75 / 74.55) & 86.45 (91.73 / 81.75) & 72.38 (88.13 / 61.41) & 67.99 (65.23 / 71.00) \\
        \midrule
        \multirow{6}{*}{\shortstack{Weakly\\Supervised\\Models}}
        & ConNet* & 66.02 (67.98 / 64.19) & 63.04 (74.55 / 55.16) & 72.04 (77.71 / 67.18) & 50.36 (63.04 / 42.73) & 60.58 (59.43 / 61.83) \\
        & MV* & 60.36 (59.06 / 61.72) & 78.44 (93.04 / 67.79) & 80.73 (83.79 / 77.88) & 73.27 (88.86 / 62.33) & 58.85 (54.17 / 64.40)\\
        & Snorkel* & 62.43 (61.62 / 63.26) & 78.44 (93.04 / 67.79) & 83.50 (91.69 / 76.65) & 73.27 (88.86 / 62.33) & 61.85 (57.44 / 66.99) \\
        & HMM* & 62.18 (66.42 / 58.45) & 66.80 (\textbf{96.79} / 51.00)  & 71.57 (\textbf{93.48} / 57.98) & 73.63 (89.30 / 62.63) & 55.67 (57.95 / 53.57) \\
        & CHMM* & 63.22 (61.93 / 64.56) & 78.74 (93.21 / 68.15)  & 83.66 (91.76 / 76.87)  & 73.26 (88.79 / 62.36) & 64.06 (59.70 / \textbf{69.09})\\
        \cmidrule{2-7}
        & \ours & \textbf{71.53} (\textbf{73.80} / \textbf{69.39}) & \textbf{82.24} (93.18 / \textbf{73.60})  &  \textbf{86.63} (89.56 / \textbf{83.88})  & \textbf{75.90} (\textbf{91.94} / \textbf{64.62}) & \textbf{64.85} (\textbf{61.26} / 68.88) \\
        \bottomrule
      \end{tabular}
      \begin{tablenotes}
        \footnotesize{
          \item * Results are from the Wrench benchmark \citep{zhang-2021-wrench}.
          All weakly supervised models are evaluated with identical data and weak annotations.
        }
      \end{tablenotes}
    \end{threeparttable}
    \label{tb:results.domains}
\end{table*}

\subsection{Main Results}
\label{subsec:main.results}

Table~\ref{tb:results.domains} shows the performance comparison of \ours and the baselines in the Wrench benchmark.
\ours outperforms all other label models, achieving \num{3.01} average \fone improvement over the strongest baselines.
In general, the increment in recall contributes to most of \ours's performance gain.
This is accredited to the well-founded emission structure, which prevents the correct annotations of inferior LFs from being neglected.
On \num{3} out of \num{5} datasets, \ours has larger recall than the best consensus.
This situation indicates that \ours is capable of predicting entities observed by \emph{no} LF, which is attributed to the well-trained token-wise transition probabilities.

Surprisingly, we find that \ours is generally superior to CHMM-FE, a model that incorporates true labels.
The reason is that the ground-truth emission matrix in CHMM-FE is not necessarily the best when used as model parameters.
Moreover, as the emission varies according to input sentences, using a fixed average value renounces the flexibility to accommodate the sentence pattern, which can be captured by \ours.

Nonetheless, a gap exists between \ours and the best consensus on CoNLL 2003 and OntoNotes 5.0.
On these datasets, even CHMM-FE fails to achieve good performance.
Apart from the difficulty these datasets bring with the increased number of entities and LFs, the reason is primarily the quality of their LFs (\cref{appsec:lf.metrics}).
The majority of LFs have low recall scores, letting the few reasonably-performed LFs dominate the training, which impedes the attention to the labels correctly observed by these LFs.

\begin{table}[t!]\small
    \caption{
      \fone scores of ablation studies.
    }
    \centering
    \begin{threeparttable}
    \begin{tabular}{c|c|c|c|c|c}
      \toprule
      & CoNLL & NCBI & BC5CDR & Laptop & OntoNotes \\
      \midrule
      \ours & \textbf{71.53} & \textbf{82.24} & \textbf{86.63} & \textbf{75.90} & 64.85 \\
      \midrule
      \multicolumn{6}{c}{Training Stages} \\
      \midrule
      SpMM S1 & 69.73 & 77.89 & 86.15 & 73.59 & 64.15 \\
      SpMM S2 & 70.79 & 79.18 & 86.04 & 74.42 & \textbf{64.92} \\
      \midrule
      \multicolumn{6}{c}{Model Components} \\
      \midrule
      Na\"ive emiss & 33.02 & 77.32 & -* & 71.77 & -* \\
      w/o $h_{nsr}$ & 58.66 & -* & 86.48 & 75.38 & 64.69 \\
      w/o SoftMax & 62.86 & 80.61 & 82.56 & 73.71 & 53.19 \\
      w/o Dirichlet & 64.70 & 81.34 & 86.05 & 73.51 & 64.64 \\
      w/o S2 & 71.18 & 81.48 & 86.02 & 73.11 & 63.90 \\
      Merge S2 \& S3 & 71.27 & 79.81 & 85.95 & 71.49 & 64.22 \\
      \bottomrule
      \end{tabular}
      \begin{tablenotes}
        \footnotesize{
          \item * The model fails training; the output labels are all ``\lbfont{O}''.
          \item ``SpMM'' represents \ours and ``S$i$'' is short for ``stage $i$''.
        }
      \end{tablenotes}
    \end{threeparttable}
    \label{tb:ablation.study}
\end{table}


\subsection{Ablation Studies}
\label{subsec:ablation.studies}

\subsubsection{Training Stages}
\label{subsec:training.stages}
Table~\ref{tb:ablation.study} shows the test performance from different training stages.
The results demonstrate that the three-stage strategy contributes to \ours's performance in general.
On BC5CDR, although stage 2 slightly weakens the model, the addon prior built in this stage is vital for stage 3's training.
On OntoNotes 5.0, the minor decrease in stage 3 is because of the discrepancy between the training and test datasets, as we observe the valid \fone increases from stage 2 to stage 3, as expected.

\subsubsection{Model Components}
\label{subsec:model.components.ablation}
To investigate the effectiveness of each model component, we include a series of ablation studies:
1)~\ours with \textbf{na\"ive emission} matrices, which substitutes the reliability expansion function \eqref{eq:gnr} by a Uniform over all non-diagonal values so that the emit-to-\lbfont{O} probabilities are not emphasized as the original model;
2)~\ours \textbf{w/o} the scaling function $h_{n,s,r}$, which is equal to substituting \eqref{eq:hnsr} by $\tilde{\A} = \hat{\A}$ or setting the exponents $n=s=1$;
3)~\ours \textbf{w/o} the \textbf{SoftMax} function defined in \eqref{eq:softmax-A};
4)~\ours \textbf{w/o} the \textbf{Dirichlet} sampling process during training;
5)~\ours \textbf{w/o stage 2} training or the weighted XOR scores;
6)~\ours that is trained with \textbf{merged stage 2 and 3}, \ie, the transition parameters are not frozen in stage 2.

From Table~\ref{tb:ablation.study}, we conclude that each component introduced in \cref{subsec:emission} contributes to the model performance.
The model does not work with the na\"ive emission, which verifies the importance of the appropriately designed emission structure \eqref{eq:fij}.
We also find that both SoftMax and $h$ are vital for \ours to predict good reliability scores.
If Dirichlet sampling is deprecated, \ours has a higher chance to be trapped by the local optima.
The 3-stage training strategy also makes the model optimization more stable.


\begin{figure}[tbp]
  \centering{
    \subfloat[\fone from $h_{n,s,r}$ with different $n$ and $s$]{
      \label{subfig:h.parameters}
      \includegraphics[height=1.2in]{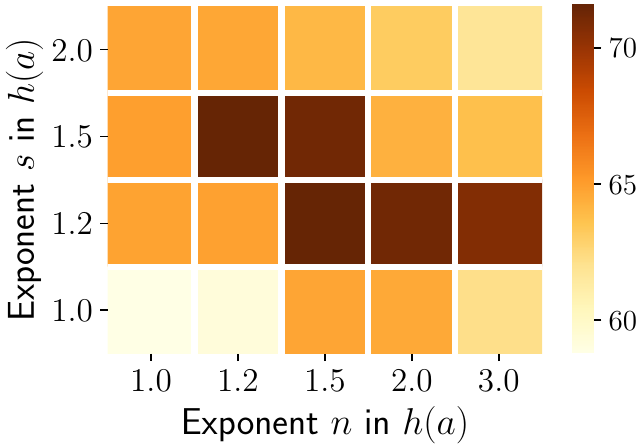}
    }
    \subfloat[$g_{n,r}$ with different $n$]{
      \label{subfig:g.parameters}
      \includegraphics[height=1.2in]{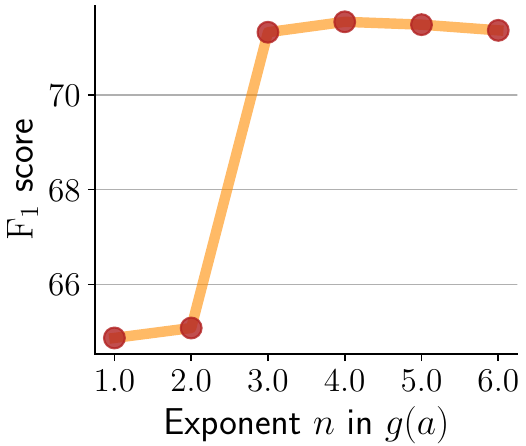}
    }
  }
  \caption{
    Hyper-parameter studies on CoNLL 2003
  }
  \label{fig:hyper.parameters}
  \Description{Ablation study}
\end{figure}

\begin{figure*}[!t]
  \centering
  \subfloat[Emissions of LF BTC\label{subfig:emiss.btc}]{%
    \includegraphics[width=0.5\textwidth]{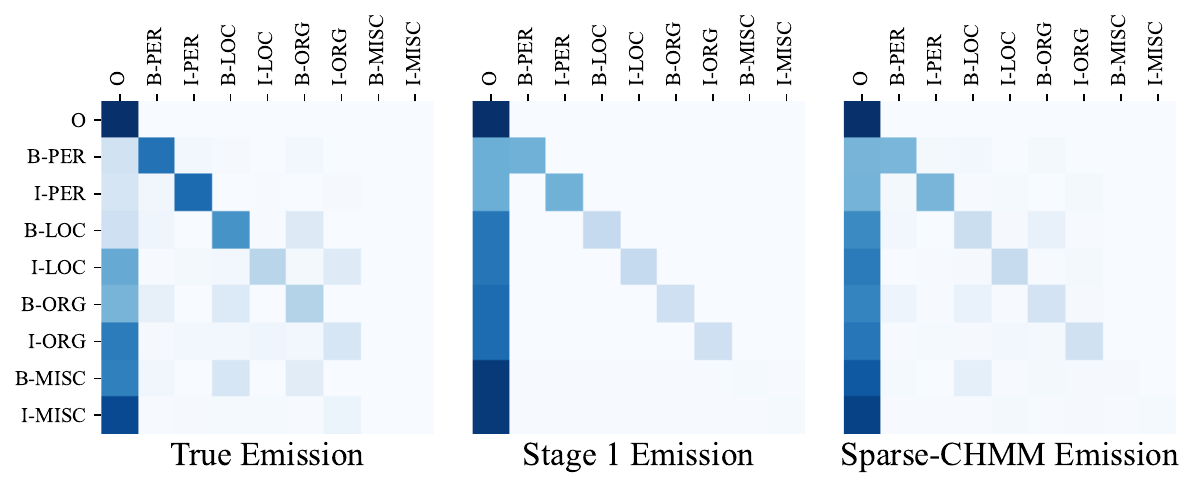}}
  \subfloat[Emissions of LF crunchbase_uncased\label{subfig:emiss.cb}]{%
    \includegraphics[width=0.5\textwidth]{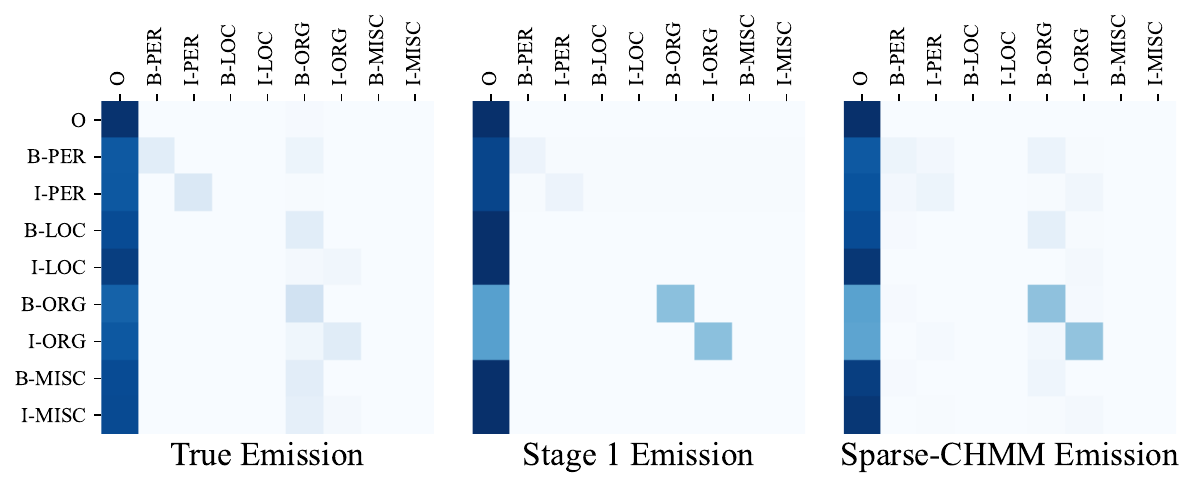}}
  \caption{Emission probabilities of two LFs on CoNLL 2003.}\label{fig:emiss.matrix.case.study}
  \Description{Case study}
\end{figure*}

\subsubsection{Hyper-Parameters}
Figure~\ref{fig:hyper.parameters} presents the impact of different exponential terms of functions $h_{n,s,r}$ and $g_{n,r}$ to the model performance.
The exponents help the element values of the base prior $\bLambda$ to rest within the desired range instead of leaving them arbitrary.
This is essential for \ours to generate good base emission patterns and calculate accurate WXOR scores.

\subsubsection{Model Efficiency}
Table~\ref{tb:training.time} supports that compared with CHMM, \ours is advanced in training efficiency.
The advantage is more prominent with an increasing number of LFs and labels, aligning with our analysis in \cref{subsec:training.complexity}.
Since \ours is a simpler model than CHMM, it not only is faster to train per epoch, but also requires less epochs to be fully optimized.
Thus, \ours is more efficient than CHMM from both perspectives.

\begin{table}[htbp]\small
    \caption{Training efficiency as ``second/epoch (\# epoch)$^*$''.}
    \centering
    \begin{threeparttable}
    \begin{tabular}{c|c|c|c|c|c}
      \toprule
      & CoNLL & NCBI & BC5CDR & Laptop & OntoNotes \\
      \midrule
      CHMM & 25.46 (50) & 5.96 (55) & 8.54 (72) & 2.60 (56) & 729.38 (28) \\
      \midrule
      SpMM S1 & 9.64 (30) & 4.66 (42) & 4.02 (87) & 2.06 (25) & 277.60 (3) \\
      SpMM S2 & 8.36 (5) & 3.92 (3) & 3.36 (5) & 1.82 (6) & 294.23 (2) \\
      SpMM S3 & 8.50 (4) & 4.04 (5) & 3.54 (9) & 1.84 (10) & 250.16 (2) \\
      \bottomrule
      \end{tabular}
      \begin{tablenotes}
        \footnotesize{
          \item[] $^*$The number of epochs required to get similar results as presented in Table~\ref{tb:results.domains}.
          \item The batch size is ajusted such that the models consume similar computation resources during training and inference.
          \item ``SpMM'' represents \ours and ``S$i$'' is short for ``stage $i$''.
        }
      \end{tablenotes}
    \end{threeparttable}
    \label{tb:training.time}
\end{table}

\subsection{Case Studies}
\label{subsec:case.studies}
Figure~\ref{fig:emiss.matrix.case.study} illustrates the averaged emission matrix from \ours and compares it with the ground-truth emissions.
We can see that the model focuses on the diagonal and emit-to-\lbfont{O} (\ie, first-column) values in the first stage, and refines the emission matrix by adding the addon prior $\bDelta$ to include the prominent off-diagonal values into the Dirichlet parameters.
Comparing the diagonal values of the true and predicted emission matrices, we find that \ours fits the LF reliability scores well without using any clean labeled data.
The reliability scores can also facilitate other tasks such as LF design and evaluation.

\begin{figure}[tbp]
  \centering{
    \subfloat[Predicted LF reliabilities \vs LF entity \fone]{%
      \label{subfig:reliability.scatter}%
      \includegraphics[height=1.6in]{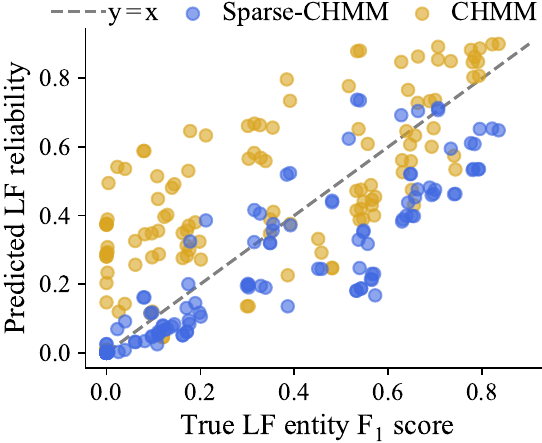}%
    }\hfil
    \subfloat[Correlation coefficients]{%
      \label{subfig:correlation.coefficient}%
      \includegraphics[height=1.6in]{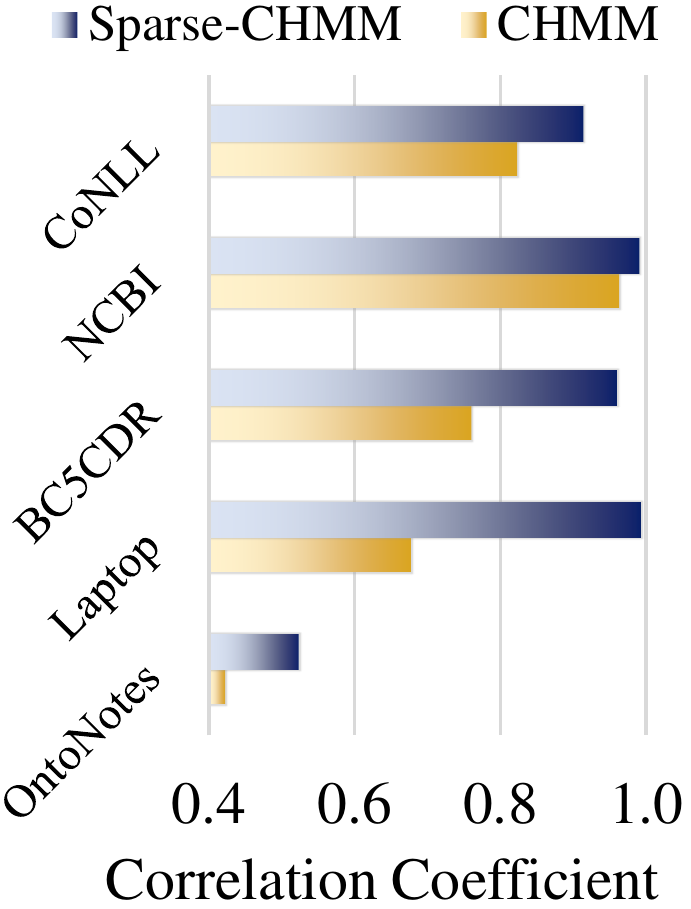}%
    }
  }
  \caption{
    Illustration of the correlation between the predicted LF reliability scores and the true LF \fone scores.
    \protect\subref{fig:lf.examples} shows the results from the CoNLL 2003 dataset.
  }
  \label{fig:lf.performance.case.study}
  \Description{Case study}
\end{figure}

We further investigate the correlation between the predicted LF reliabilities and the entity \fone scores of LFs.
Figure~\ref{subfig:reliability.scatter} illustrates correlation, and Figure~\ref{subfig:correlation.coefficient} measures the correlation quantitatively.
We observe that the predicted reliabilities have a strong correlation with \fone scores, reaching almost $1.0$ correlation coefficients on $3$ datasets.
Compared with CHMM, \ours performs better in identifying unreliable LFs, which is critical for ruling out the incorrect observations.

\section{Related Works}
\label{sec:related.works}

Weak supervision is a widely-investigated approach for natural language processing tasks such as sentence classification \cite{Ratner-2016-data-programming, Ratner-2017-Snorkel, ren-etal-2020-denoising} and NER \cite{nguyen-etal-2017-aggregating, shang-etal-2018-learning, Safranchik-etal-2020-weakly, lison-etal-2020-named, lan-etal-2020-learning, lison-etal-2021-skweak, parker-yu-2021-named, li-etal-2021-bertifying}.
It substitutes the laborious manual annotation process with multiple simple labeling functions, relieving the human effort to annotate large training data.
However, the LFs are generally noisy and controversial, which makes their utilization challenging.
Consequently, the primary track of weak supervision focuses on designing a robust label model that aggregates the weak annotations by de-noising them and resolving the conflict \cite{Ratner-2016-data-programming, Ratner-2017-Snorkel, Ratner-2019-training-complex, ren-etal-2020-denoising, nguyen-etal-2017-aggregating, shang-etal-2018-learning, Safranchik-etal-2020-weakly, lison-etal-2020-named, lan-etal-2020-learning, lison-etal-2021-skweak, parker-yu-2021-named, li-etal-2021-bertifying}.
The hidden Markov model with multiple independent observation sequences is principled and popular for this task \citep{nguyen-etal-2017-aggregating, Safranchik-etal-2020-weakly, lison-etal-2020-named,lison-etal-2021-skweak, parker-yu-2021-named, li-etal-2021-bertifying}.
Transferring from sequence classification to sequence labeling (\aka, token classification), it extends the na\"ive Bayes generative model proposed by \citet{Ratner-2016-data-programming} by incorporating the chronological dependency relationship through the transition matrix.
The weak labels from different LFs are considered as multiple sets of observations that are conditionally independent given the latent variables.
Among these works, \citet{nguyen-etal-2017-aggregating} and \citet{lison-etal-2020-named, lison-etal-2021-skweak} use conventional HMMs.
\citet{Safranchik-etal-2020-weakly} design the linked HMM, which uses linking rules to model whether consecutive tokens belong to the same entity.
Other works \citep{li-etal-2021-bertifying, parker-yu-2021-named} improve HMM's context representation with BERT embeddings and alleviate the Markov assumption.

In the data programming framework, people also train \emph{end models}, which are deep supervised neural networks, with the label model outputs in seek of refining the results with the power of deep networks.
One exception is the Consensus Network \citep{lan-etal-2020-learning} that trains the label model and the end model jointly within a two-stage framework \citep{zhang-2021-wrench}.
Another is ALT \citep{li-etal-2021-bertifying}, which treats the end model as an additional LF and optimizes it alternately with the label model.
\citet{zhang-2021-wrench}, however, show that the end model does not necessarily outperform the label model. 
Therefore, we focus on the label models here and leave end models to future works.

Our work is also related to the neuralized graphical models.
Many efforts seek to inject neural networks into graphical models, especially for variational inference, such as the variational autoencoder (VAE) \citep{Kingma-2014-VAE}.
Other works include \citep{tran-etal-2016-unsupervised}, which neuralizes the HMM with one observation set for the unsupervised part-of-speech tagging.
\citet{Dai-2017-recurrent} and \citet{Liu-2018-structured-inference} incorporate recurrent units into the hidden semi-Markov model to segment and label high-dimensional time series;
\citet{wiseman-etal-2018-learning} learn discrete template structures for conditional text generation also with neuralized graphical models.
CHMM and DWS also fall into this category by predicting graphical models components through NNs.



\section{Conclusion}
\label{sec:conclusion}

We presented \ours, a label model that aggregates the weak annotations from multiple noisy NER labeling functions.
In contrast to CHMM, \ours constructs the sentence-level sparse emission probabilities from the LF reliabilities predicted using the BERT sentence embeddings.
The off-diagonal emission elements are further augmented by the weighted XOR scores, which estimate the possibilities of an LF observing incorrect entity labels.
Wrapped in a three-stage training process with Dirichlet sampling, \ours outperforms all baseline label models on five representative datasets with different statistics and attributes.
In addition, the approximated LF reliability scores strongly correlate with the true LF performance, making it eligible to contribute to other tasks such as automated LF generation and evaluation.
In the future, we will consider leveraging prior heuristic knowledge to strengthen \ours's ability to distill clean NER labels from noisy LF annotations.
In addition, we also plan to extend this technique to other sequence labeling tasks.

\begin{acks}
  This work was supported in part by ONR MURI N00014-17-1-2656, NSF IIS-2008334, IIS-2106961, CAREER IIS-2144338, and Kolon Industries.
\end{acks}

\bibliographystyle{ACM-Reference-Format}
\bibliography{reference.st}

\appendix

\section{Training Details}
\label{appsec:objective}

\subsection{Training Objective}
\label{appsubsec:objective}

In this section, we focus on the computation of the expected complete data log likelihood $Q$ defined in \eqref{eq:em.q} as well as the training objective.
The derivation largely aligns with \citep{li-etal-2021-bertifying}, so we will skip some trivial steps and explanations.

By inserting \eqref{eq:lc} into \eqref{eq:em.q} and specifying the expectation of $z$, we can write \eqref{eq:em.q} as:
\begin{equation}
    \label{eq:q.calculate}
    \begin{aligned}
        Q&(\btht, \btht^{\rm old}) = \sum_{i=1}^{L} p(z^{(0)} = i| \x^{(1:T)}, \e^{(0:T)}) \log p(z^{(0)}=i) + \\
        & \sum_{t=1}^{T}\sum_{i=1}^{L} \sum_{j=1}^{L} p(z^{(t-1)}=i, z^{(t)}=j | \x^{(1:T)}, \e^{(0:T)}) \log \Psi^{(t)}_{i,j} +\\
        & \sum_{t=1}^{T}\sum_{i=1}^{L} p(z^{(t)} = i| \x^{(1:T)}, \e^{(0:T)}) \log \varphi^{(t)}_{i},
    \end{aligned}
\end{equation}
where $\bPsi$ is the transition matrix and $\varphi_i^{(t)}$ is defined in \eqref{eq:varphi}.
$p(z^{(0)})$ is the probability of the initial hidden state without any corresponding observations.
As we can predict the token-wise transition matrix from the embeddings, we can simply set it to Uniform or, as \citet{li-etal-2021-bertifying} proposed, set $p(z^{(0)}=1)$ to $1$ and $p(z^{(0)}=i), \forall i\in 2:L$ to $0$.

To calculate \eqref{eq:q.calculate}, we define the smoothed marginal $\bgamma^{(t)}\in [0,1]^{L}$ as:
\begin{equation*}
    \gamma^{(t)}_i \triangleq p(z^{(t)} = i| \x^{(1:T)}, \e^{(0:T)}),
\end{equation*}
and the expected number of transitions $\bxi^{(t)} \in [0,1]^{L\times L}$ as:
\begin{equation*}
    \xi^{(t)}_{i,j} \triangleq p(z^{(t-1)}=i, z^{(t)}=j | \x^{(1:T)}, \e^{(0:T)}).
\end{equation*}
These two variables are acquired using the \emph{forward-backward} algorithm.

First, we define the filtered marginal $\balpha \in [0,1]^L$ as:
\begin{equation*}
    \alpha^{(t)}_i \triangleq p(z^{(t)}=i|\x^{(1:t)}, \e^{(0:T)}),
\end{equation*}
and the conditional future evidence $\bbeta \in [0,1]^{L}$ as:
\begin{equation*}
    \beta^{(t)}_i \triangleq p(\x^{(t+1:T)}|z^{(t)} = i, \e^{(0:T)}).
\end{equation*}

In the forward pass, $\alpha^{(t)}_i$ is computed iteratively:
\begin{equation*}
  \begin{aligned}
    \alpha^{(t)}_i &\propto p(\x^{(t)}|z^{(t)}=i, \e^{(0)}) p(z^{(t)}=i|\x^{(1:t-1)}, \e^{(0:t)}) \\
    &= \sum_{j=1}^L \varphi^{(t)}_i \Psi^{(t)}_{j,i} \alpha^{(t-1)}_j,
  \end{aligned}
\end{equation*}
which can be written in the matrix form:
\begin{equation*}
    \balpha^{(t)} \propto \bphi^{(t)} \odot ({\bPsi^{(t)}}^{\sf T}\balpha^{(t-1)}),
\end{equation*}
where $\odot$ is the element-wise product.
We initialize $\balpha$ with $\alpha^{(0)}_l = p(z^{(0)}=l), \forall l\in 1:L$ since we have no observation at time step $0$.

Similarly, we do the backward pass and compute $\bbeta$:
\begin{equation*}
    \begin{aligned}
      \beta^{(t-1)}_i &= \sum_{j=1}^L p(z^{(t)}=j, \x^{(t)}, \x^{(t+1:T)}|z^{(t-1)}=i, \e^{(0,t:T)}) \\
      &= \sum_{j=1}^L \beta^{(t)}_j \varphi^{(t)}_j \Psi^{(t)}_{i,j}.
    \end{aligned}
\end{equation*}
In the matrix form, it becomes:
\begin{equation*}
    \bm{\beta}^{(t-1)} = \bPsi^{(t)}(\bphi^{(t)} \odot \bm{\beta}^{(t)}),
\end{equation*}
with base case:
\begin{equation*}
    \beta^{(T)}_i = p(\x^{(T+1:T)}|z^{(T)}=i) = 1, \forall i \in 1:L.
\end{equation*}

With $\balpha$ and $\bbeta$ calculated, $\gamma^{(t)}_i$ and $\xi^{(t)}_{i,j}$ can be written as:
\begin{gather*}
    \begin{aligned}
        \gamma^{(t)}_i &\propto p(z^{(t)}=i|\x^{(1:t)}, \e^{0:t}) p(\x^{(t+1:T)}|z^{(t)} = i, \e^{(0, t+1:T)}) \\
        &= \alpha^{(t)}_i \beta^{(t)}_i,
    \end{aligned} \\
    \begin{aligned}
        \xi^{(t)}_{i,j} & \propto p(z^{(t-1)}=i|\x^{(1:t-1)} \e^{(0:t-1)}) p(\x^{(t)}|z^{(t)}=j, \e^{(0)}) \\
        & \quad \ p(\x^{(t+1:T)}|z^{(t)}=j, \e^{(0, t+1:T)})p(z^{(t)}=j|z^{(t-1)}=i, \e^{(t)})\\
        &= \alpha^{(t-1)}_i\varphi^{(t)}_j\beta^{(t)}_j\Psi^{(t)}_{i,j}.
      \end{aligned}
\end{gather*}
Written in the matrix form, they become:
\begin{gather*}
    \bm{\gamma}^{(t)} \propto \balpha^{(t)} \odot \bm{\beta}^{(t)}, \\
    \bm{\xi}^{(t)} \propto \bPsi^{(t)} \odot (\balpha^{(t-1)}(\bphi^{(t)} \odot \bm{\beta}^{(t)})^{\sf T}).
\end{gather*}

Eventually, we insert $\bgamma$ and $\bxi$ into \eqref{eq:q.calculate} to compute the value of $Q$.
The training objective is to maximize $Q$, which can be readily done using the gradient ascend.
Please refer to \citep{li-etal-2021-bertifying} for more details.

\subsection{Inference}
\label{appsubsec:inference}

Same as \citet{li-etal-2021-bertifying}, we use the Viterbi algorithm to find the sequence of latent variables $\hat{\z}^{(1:T)}$ that maximize the posterior:
\begin{equation*}
    \hat{\z}^{(1:T)} = \argmax_{\z^{(1:T)}}p(\z^{(1:T)} | \x^{(1:T)}, \e^{(0:T)}).
\end{equation*}
This sequence of latent variables $\hat{\z}^{(1:T)}$ is considered as \ours's approximation of the true labels $\y$.

\section{Labeling Function Metrics}
\label{appsec:lf.metrics}

Table~\ref{tb:lf.performance.conll.2003}--\ref{tb:lf.performance.ontonotes} present the performance of each labeling function on the test set.

\begin{table}[htbp]\small
    \caption{LF performance on CoNLL 2003.}
    \centering
    \begin{tabular}{c|c|c|c}
    \toprule
    LF name & Precision & Recall & \fone \\
    \midrule
    BTC &   67.26 &   44.56 &   53.61 \\
    core_web_md &   70.52 &   58.27 &   63.81 \\
    crunchbase_cased &   37.76 &   6.69 &   11.37 \\
    crunchbase_uncased &   32.75 &   7.31 &   11.96 \\
    full_name_detector &   84.63 &   11.60 &   20.40 \\
    geo_cased &   67.99 &   16.63 &   26.72 \\
    geo_uncased &   64.35 &   20.20 &   30.75 \\
    misc_detector &   85.26 &   20.68 &   33.29 \\
    multitoken_crunchbase_cased &   72.73 &   3.40 &   6.50 \\
    multitoken_crunchbase_uncased &   71.22 &   3.51 &   6.68 \\
    multitoken_geo_cased &   72.06 &   1.74 &   3.39 \\
    multitoken_geo_uncased &   66.83 &   2.35 &   4.55 \\
    multitoken_wiki_cased &   94.36 &   16.01 &   27.37 \\
    multitoken_wiki_uncased &   90.55 &   16.63 &   28.09 \\
    wiki_cased &   75.62 &   35.48 &   48.30 \\
    wiki_uncased &   71.50 &   38.95 &   50.43 \\
    \bottomrule
    \end{tabular}
    \label{tb:lf.performance.conll.2003}
\end{table}

\begin{table}[htbp]\small
    \caption{LF performance on NCBI-Disease.}
    \centering
    \begin{tabular}{c|c|c|c}
    \toprule
    LF name & Precision & Recall & \fone \\
    \midrule
    tag-CoreDictionaryUncased &   80.99 &   41.39 &   54.79 \\ 
    tag-CoreDictionaryExact &   80.69 &   17.21 &   28.37 \\ 
    tag-CancerLike &   34.88 &   1.58 &   3.03 \\ 
    tag-BodyTerms &   68.52 &   3.91 &   7.39 \\ 
    link-ExtractedPhrase &   96.88 &   36.11 &   52.62 \\ 
    \bottomrule
    \end{tabular}
    \label{tb:lf.performance.ncbi.disease}
\end{table}

\begin{table}[htbp]\small
    \caption{LF performance on BC5CDR.}
    \centering
    \begin{tabular}{c|c|c|c}
    \toprule
    LF name & Precision & Recall & \fone \\
    \midrule
    tag-DictCore-Chemical &   93.24 &   29.68 &   45.03  \\
    tag-DictCore-Chemical-Exact &   89.55 &   3.26 &   6.29  \\
    tag-DictCore-Disease &   84.19 &   26.91 &   40.78  \\
    tag-DictCore-Disease-Exact &   81.40 &   1.08 &   2.13  \\
    tag-Organic Chemical &   94.06 &   30.17 &   45.68  \\
    tag-Antibiotic &   97.88 &   2.38 &   4.64  \\
    tag-Disease or Syndrome &   79.01 &   11.81 &   20.55  \\
    link-PostHyphen &   86.24 &   7.93 &   14.53  \\
    link-ExtractedPhrase &   87.21 &   17.88 &   29.68  \\
    \bottomrule
    \end{tabular}
    \label{tb:lf.performance.bc5cdr}
\end{table}

\begin{table}[htbp]\small
    \caption{LF performance on LaptopReview.}
    \centering
    \begin{tabular}{c|c|c|c}
    \toprule
    LF name & Precision & Recall & \fone \\
    \midrule
    tag-CoreDictionary &   72.63 &   51.61 &   60.34  \\
    link-ExtractedPhrase &   97.46 &   29.40 &   45.18  \\
    link-ConsecutiveCapitals &   35.29 &   0.92 &   1.79  \\
    \bottomrule
    \end{tabular}
    \label{tb:lf.performance.laptop.review}
\end{table}

\begin{table}[htbp]\small
    \caption{LF performance on OntoNotes 5.0.}
    \centering
    \begin{tabular}{c|c|c|c}
    \toprule
    LF name & Precision & Recall & \fone \\
    \midrule
    Core_Keywords &   51.25 &   8.01 &   13.86  \\
    Regex_Patterns &   75.08 &   3.44 &   6.58  \\
    Numeric_Patterns &   60.58 &   10.71 &   18.21  \\
    wiki_fine &   77.24 &   53.15 &   62.97  \\
    money_detector &   47.37 &   1.40 &   2.72  \\
    date_detector &   66.74 &   4.34 &   8.15  \\
    number_detector &   39.71 &   5.70 &   9.97  \\
    company_type_detector &   65.14 &   1.43 &   2.80  \\
    full_name_detector &   53.62 &   4.54 &   8.37  \\
    misc_detector &   61.75 &   10.84 &   18.44  \\
    crunchbase_cased &   23.07 &   4.80 &   7.94  \\
    crunchbase_uncased &   22.63 &   4.84 &   7.97  \\
    geo_cased &   62.87 &   9.03 &   15.79  \\
    geo_uncased &   62.73 &   9.05 &   15.82  \\
    Multitoken_wiki &   87.38 &   13.06 &   22.72  \\
    wiki_cased &   48.61 &   15.62 &   23.64  \\
    wiki_uncased &   48.24 &   15.68 &   23.67  \\
    \bottomrule
    \end{tabular}
    \label{tb:lf.performance.ontonotes}
\end{table}

\begin{table}[htbp]\small
    \caption{
    Hyper-parameters.
    }
    \centering
    \begin{threeparttable}
        \begin{tabular}{cc|c|c|c|c|c}
        \toprule
        & & CoNLL & NCBI & BC5CDR & Laptop & OntoNotes \\
        \midrule
        \multicolumn{7}{c}{Training hyper-parameters} \\
        \midrule
        \multicolumn{2}{c|}{Batch size} & 256 & 128 & 128 & 256 & 32 \\
        \multicolumn{2}{c|}{PT LR} & 5e-4 & 5e-4 & 5e-4 & 5e-4 & 1e-4 \\
        \multicolumn{2}{c|}{LR (S1)} & 2e-4 & 1e-3 & 1e-3 & 1e-4 & 1e-4 \\
        \multicolumn{2}{c|}{LR (S2)} & 4e-5 & 2e-4 & 2e-4 & 2e-5 & 2e-5 \\
        \multicolumn{2}{c|}{LR (S3)} & 2e-4 & 1e-3 & 1e-3 & 1e-4 & 1e-4 \\
        \multicolumn{2}{c|}{Use MV$^1$} & false & true & false & true & false \\
        \midrule
        \multicolumn{7}{c}{Model hyper-parameters} \\
        \midrule
        \multicolumn{2}{c|}{Reliab LV} & ENT & LB & ENT & LB & ENT \\
        \multicolumn{2}{c|}{$\nu^{\rm base}$} & 10 & 2 & 2 & 2 & 2 \\
        \multicolumn{2}{c|}{$\nu^{\rm expan}$} & \num{1000} & \num{1500} & \num{1500} & \num{1000} & \num{1000} \\
        \midrule
        \multirow{3}{*}{$h$}
        & $n$ & 1.2 & 1.2 & 0.9 & 0.8 & 1.1 \\
        & $s$ & 1.5 & 3 & 1.1 & 1.5 & 1 \\
        & $r$ & \multicolumn{5}{c}{$\frac{1}{K}$}   \\
        \midrule
        \multirow{2}{*}{$g$}
        & $n$ & \multicolumn{5}{c}{4} \\
        & $r$ (S1) & $\frac{1}{2L}$ & $\frac{1}{20L}$ & $\frac{1}{10L}$ & $\frac{1}{20L}$ & $\frac{1}{5L}$ \\
        & $r$ (S2,3) & $\frac{1}{20L}$ & $\frac{1}{20L}$ & $\frac{1}{10L}$ & $\frac{1}{20L}$ & $\frac{1}{5L}$ \\
        \bottomrule
        \end{tabular}
    \begin{tablenotes}
        \footnotesize{
        \item ``PT'' is ``pre-training''; ``Reliab LV'' is short for the ``reliability level''; ``LB'' and ``ENT'' indicate ``label''-level reliability (one score per label) and ``entity''-level reliability (one score per entity).
        \item ``S$i$'' represents ``stage $i$''; paramters with no specified stage remain constant for all training stages.
        \item $L$ and $K$ are the numbers of labels and LFs, respectively (\cref{sec:definition}).
        \item $h$ is defined in \eqref{eq:hnsr}; $g$ is defined in \eqref{eq:gnr}.
        \item[1] On some datasets, we add an additional majority voting LF to balance the annotations from existing LFs. The majority voting is only deployed in training but not involved in inference or evaluation.
        }
    \end{tablenotes}
    \end{threeparttable}
    \label{tb:hyperparameters}
\end{table}

\section{Hyper-parameters}
\label{appsec:parameters}

The experiments are conducted on one GeForce RTX 2080 Ti GPU.
The selected model hyper-parameters are presented in Table~\ref{tb:hyperparameters}.

To increase the training speed on the OntoNotes 5.0 dataset, the WXOR scores are calculated only on the validation set instead of the combination of training and validation sets.

\section{Function design criteria}
\label{appsec:function.criteria}

Here we introduce the design criteria for the reliability scaling function \eqref{eq:hnsr} and the reliability expansion function \eqref{eq:gnr}.
There are infinite solutions that meet the criteria.
We select the simplest polynomials for better interpretability and calculation efficiency.

\subsection{Reliability Scaling Function}

Function $f_{n,s,r}$, illustrated in Figure~\ref{fig:ha}, is 1) continuous, smooth and monotonic; 2) passing through coordinates $(0,0)$ and $(1,1)$; and 3) having zero gradient at $(0,0)$ and $(1,1)$.
It is designed such that its shape is controllable without making the function complicated.

\subsection{Reliability Expansion Function}

As shown in Figure~\ref{fig:ga}, $g_{n,r}$ is 1) continuous, smooth and monotonic; 2) passing through $(0,1)$ and $(1,0)$; and 3) $\nabla_{a}g(a)|_{a=0} = \frac{1}{L-1}$.
We want the emission from non-\lbfont{O} latent state to non-\lbfont{O} observation $\bLambda_{k,i,j\geq 2}$ to be close to Uniform when $\tilde{A}_{k,i}$ is small, which indicates that LF $k$ may equally observe anything when the latent label is $i$.
This builds the constraint of $\nabla_{a}\frac{1-a-g(a)}{L-2}|_{a=0}=\nabla_{a}a|_{a=0}=1$, which leads to the third feature defined above.
The hyperparameter $r$ controls the threshold where we trust LF $k$ enough to stop increasing the off-diagonal emissions.

\end{document}